\documentclass[letterpaper, 10 pt, conference]{ieeeconf}  % Comment this line out if you need a4paper

\IEEEoverridecommandlockouts                              % This command is only needed if 
\usepackage{graphics} % for pdf, bitmapped graphics files
\usepackage{epsfig} % for postscript graphics files
\usepackage{amsmath} % assumes amsmath package installed
\usepackage{amssymb}  % assumes amsmath package installed

\usepackage{cite}
\usepackage{url}
\usepackage[margin=0.75in]{geometry}
\usepackage{overpic}
\usepackage{xcolor}
\usepackage{tikz}

\title{\LARGE \bf
Industrial Application of 6D Pose Estimation for Robotic Manipulation in Automotive Internal Logistics
}

\author{Philipp Quentin$^{1,2}$, Dino Knoll$^{1}$ and Daniel Goehring$^{2}$% <-this % stops a space
\thanks{\copyright 2023 IEEE.  Personal use of this material is permitted.  Permission from IEEE must be obtained for all other uses, in any current or future media, including reprinting/republishing this material for advertising or promotional purposes, creating new collective works, for resale or redistribution to servers or lists, or reuse of any copyrighted component of this work in other works.}%
\thanks{$^{1}$BMW Group, Munich, Germany}%
\thanks{$^{2}$Freie Universitaet Berlin, Dahlem Center for Machine Learning and Robotics, Berlin, Germany }%
}

\begin{document}

\maketitle
\thispagestyle{empty}
\pagestyle{empty}

%%%%%%%%%%%%%%%%%%%%%%%%%%%%%%%%%%%%%%%%%%%%%%%%%%%%%%%%%%%%%%%%%%%%%%%%%%%%%%%%
\begin{abstract}

Despite the advances in robotics a large proportion of the of parts handling tasks in the automotive industry's internal logistics are not automated but still performed by humans. A key component to competitively automate these processes is a 6D pose estimation that can handle a large number of different parts, is adaptable to new parts with little manual effort, and is sufficiently accurate and robust with respect to industry requirements. In this context, the question arises as to the current status quo with respect to these measures. To address this
we built a representative 6D pose estimation pipeline with state-of-the-art components from economically scalable real to synthetic data generation to pose estimators and evaluated it on automotive parts with regards to a realistic sequencing process. We found that using the data generation approaches, the performance of the trained 6D pose estimators are promising, but do not meet industry requirements. We reveal that the reason for this is the inability of the estimators to provide reliable uncertainties for their poses, rather than the ability of to provide sufficiently accurate poses. In this context we further analyzed how RGB- and RGB-D-based approaches compare against this background and show that they are differently vulnerable to the domain gap induced by synthetic data.
 
\end{abstract}

%%%%%%%%%%%%%%%%%%%%%%%%%%%%%%%%%%%%%%%%%%%%%%%%%%%%%%%%%%%%%%%%%%%%%%%%%%%%%%%%

\section{INTRODUCTION}

Despite the current advances in robotics and artificial intelligence a large proportion of processes in the internal logistics of the automotive industry are still carried out by humans.
This especially concerns, but is not limited to, the process of parts sequencing for the assembly line. The essential cognitive task which is performed by humans to accomplish this process is the visual recognition that enables the picking and placing of automotive parts. The requirements for a 6D pose estimation system to competitively enable a robot to perform these tasks instead, are exceptionally high. On the one hand, this is due to the high number of different parts combined with -in the robotic sense- highly unstructured environments. On the other hand, this results from the need for short cycle times combined with near-optimal availability that leaves no room for robot errors, such as robotic crashes. This requires from the 6D pose estimation system to not only be fast, effortless maintainable and scaleable to this large number of different parts, but also to be robust against occlusions and erroneous sensor data.

Against this background the recent advances of neural networks in the field of  6D pose estimation, are promising to solve the above challenges. Current leading approaches are nowadays mainly based on neural networks and outperform traditional approaches \cite{Hodan2020}.
Together with the developed frameworks  to easily generate real-world \cite{2018_Marion_ICRA} and theoretically unlimited synthetic \cite{Morrical2020} training data, this may have the potential to competitively enable the automation of the internal logistics. Therefore the question to current state  of a  6D pose estimation pipeline applied the internal logistics in industry arises.
\begin{figure}[t!]
      \centering
      \setlength{\fboxsep}{0pt}
  \setlength{\fboxrule}{0pt}
      \framebox{\parbox{3in}{\centerline{\includegraphics[scale=0.5]{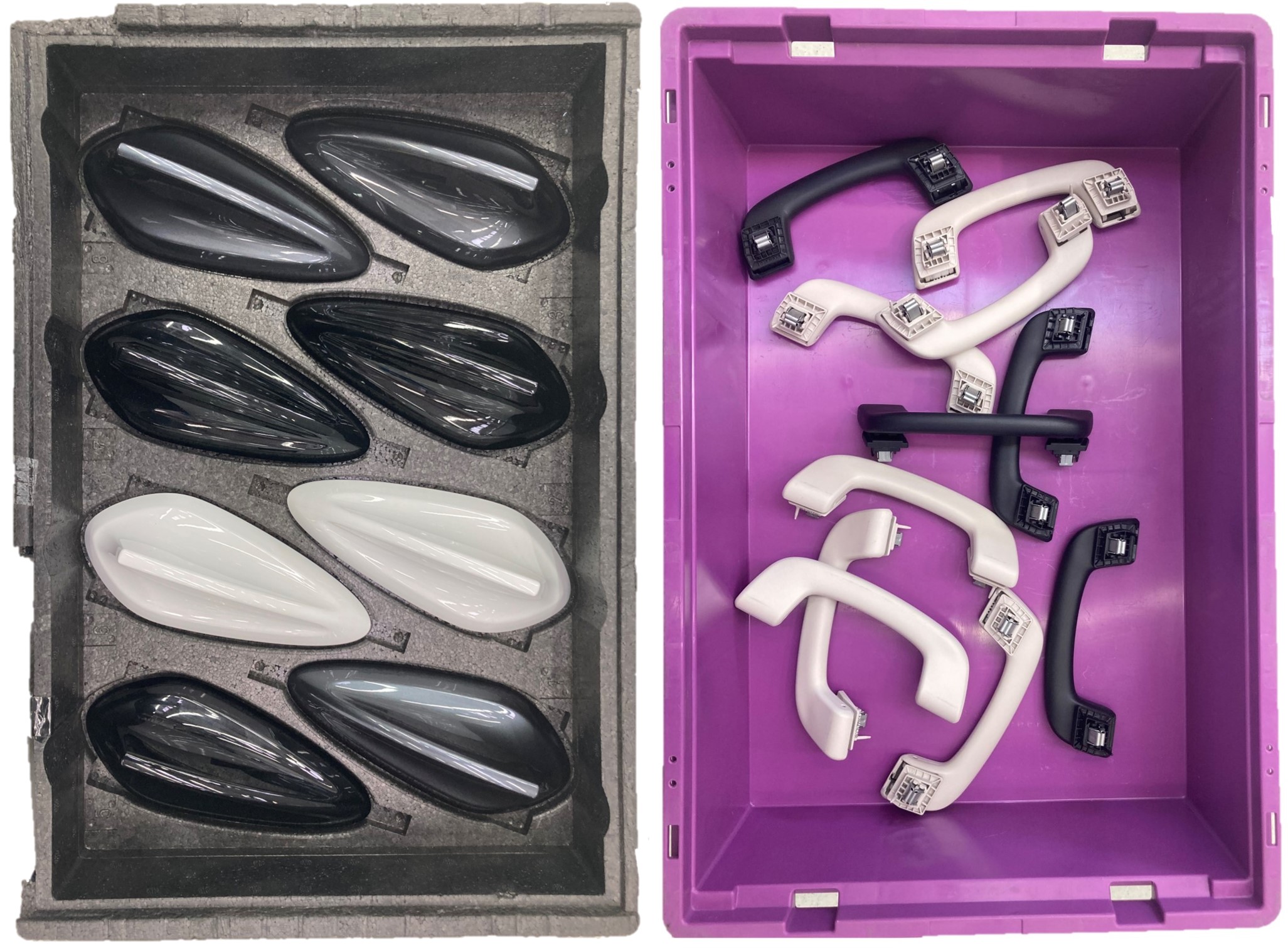}}
}}
      
      \caption{Automotive antenna covers (left) and interior handles (right) in storage containers, that enforce structured and unstructured positions.}
      \label{Fig:parts}
\end{figure}
Despite the various recent works in the fields of data generation and 6D pose estimation, there is to the best of our knowledge, no work which applies a full integrative pipeline of these blocks with such a focus. 

Therefore we aim to deepen the insights on this topic  with  building blocks from data generation to 6D pose estimation on exemplary automotive parts. To this end, we build a pipeline consisting out of three main state-of-the-art building blocks. One block for real-world data generation which implements LabelFusion \cite{2018_Marion_ICRA}, one block for synthetic data generation which applies NVISII \cite{Morrical2020} and one block for the 6D pose estimation itself. The 6D pose estimation block implements the state-of-the-art RGB-based approach GDR-Net \cite{Wang_2021_CVPR} and the RGB-D-based approach DenseFusion \cite{2019_Wang_CVPR}. We applied the pipeline to two exemplary automotive parts shown in Fig. \ref{Fig:parts}, namely: antenna covers (antennas), which are supposed to represent rather difficult parts due to their their geometric and visual properties, and interior handles (handles), which are supposed to represent rather simple parts. To  evaluate the pipeline with respect to industry requirements, we conducted experiments on the generated data sets together with robotic grasp and place tests on a representative sequencing process.
 
We found that using the data generation approaches, the performance of the trained 6D pose estimators is promising in terms of scalability, but does not meet industry requirements in terms of robustness. We show that the reason for this is the inability of the estimators to provide a reliable uncertainty measure, rather than the ability to provide sufficiently accurate poses. Furthermore, we observe that in the case of the RGB-D-based estimator, despite its reliance on neural networks, it is not robust to slightly erroneous depth data, and we infer that the circumstance of not having depth data in the case of the RGB-based estimator significantly amplifies the domain gap on the distance estimate.

\section{RELATED WORK}\label{sec:rel_work}
To the best of our knowledge, there is no direct comparable work, which evaluates a  pipeline from data generation to 6D pose estimation on automotive parts in terms of feasibility and scalability with respect to industry requirements. Therefore, we discuss the works which are directly relevant for the building blocks of our pipeline. 

\subsection{Real-World Data Generation}

The majority of the work towards methods for real-world data generation is contained in the publications of the several established data sets \cite{T-LESS, ITODD, 2018_Xiang_RSS, Homebrewed, Hodan2018}, which were created to evaluate and benchmark 6D pose estimators. A core approach among these works is to arrange the target objects in a static scene. Then either the sensor setup \cite{T-LESS, Homebrewed, 2018_Xiang_RSS} or the static scene \cite{ITODD} is moved to generate samples from different viewpoints. Subsequently, the 6D poses of the target objects are either annotated in one sample \cite{2018_Xiang_RSS, Hodan2018, ITODD} or in a 3D reconstruction \cite{Homebrewed, T-LESS} of the scene and transferred to all samples via estimation, geometric relations or by a combination of both. This strategy to transfer one annotation to all samples in a static scene, enables the annotation of large datasets with only little human effort.     

However, these works focus on the properties of the data sets and the evaluation of 6D pose estimators and less on their data generating methods. As a consequence they often do not contain detailed descriptions of these methods and the source code is not published or not completely open source as in case of \cite{Homebrewed}. In contrast to that  LabelFusion \cite{2018_Marion_ICRA} is open source and targets the data generation for 6D pose estimation in terms of practicability and scalability. It also follows the approach of static scenes in combination with a 3D reconstruction. As an improvement to \cite{Homebrewed, T-LESS} it further does not require markers in the scene, which enables a high freedom for scene generation.  Therefore, we decided for LabelFusion as our pipeline building block for real-world data generation. 

\subsection{Synthetic Data Generation}\label{sec:rel_synth}

The main challenge of synthetic data generation is to effectively bridge the domain gap so that 6D pose estimators trained on synthetic data perform equally well on real-world data. To address this challenge two complementary approaches have emerged. One is domain randomization \cite{2017_Tobin_IROS}, which aims to strongly randomize synthetic data, such as textures, backgrounds, lighting conditions, etc., so that real-world data simply represents another of these variations to the trained models. The other is to generate photorealistic data, so that the deviation of the synthetic from the real domain becomes minimal.

In this context, \cite{Kehl2017, 2018_Sundermeyer_ECCV, 2021_Hofer_ICIP} pursued the so-called "render \& paste" approach \cite{Hodan2020}, which renders objects in combination with domain randomization techniques in front of random real photographs, achieving promising results. The main difference between these works is that \cite{Kehl2017,2018_Sundermeyer_ECCV} used a rasterization-based renderer (RBR) and \cite{2021_Hofer_ICIP} used a physically-based renderer (PBR). The difference is that a PBR can accurately simulate the path of light through ray tracing, while a RBR approximates the path of light through rasterization. In consequence, a PBR enables physically-plausible domain randomization techniques, taking into account material properties and light interactions between objects, whereas a RBR is limited to approximated light effects and image-based domain randomization as for example randomized image contrasts \cite{2020_Denninger_RSS}. Against this background the BOP challenge of 2020 \cite{Hodan2020} showed that domain randomized PBR images of objects simulated in an 3D cube, representing lightweight photorealistic scenes, improved the results on 6D pose estimation compared to RBR "paste \& render" images by large margins. 
Related to that DOPE \cite{DOPE}, which was entirely trained on PBR "paste \& render" images and fully PBR photorealistic indoor and outdoor scenes, achieved on par results on the established real-world YCB-V data set \cite{2018_Xiang_RSS} with 6D pose estimators trained on real data. It was shown that both of these synthetic data generating approaches were essential to reach these results.

With regard to this, the domain randomized PBR approach of the BOP challenge to generate lightweight photorealistic scenes is a good compromise in terms of scalability, as it improves on the "paste \& render" approach but does not go along with the high effort to generate full photorealistic scenes. Therefore, we explore further on this approach.

\subsection{6D Pose Estimation}

6D pose estimation is nowadays mostly lead by deep learning based estimators. Among these, the current leading ones can be subdivided into those  rely solely on RGB images, or additionally use depth information from an RGB-D camera.\cite{2023_Sundermeyer_CVPR}

\subsubsection{RGB-based}
Deep learning based RGB 6D pose estimators can be divided into direct, correspondence, or implicit representation based estimators. 
Direct estimators \cite{2018_Xiang_RSS, 2020_Labbe_ECCV} take the RGB image as an input and directly estimate a fixed 6D pose representation via a CNN architecture. A disadvantage of these methods is that they tend to suffer from difficulties during the learning process in case of existent pose ambiguities \cite{2018_Sundermeyer_ECCV}. 
Correspondence based methods estimate the corresponding coordinates of either sparse \cite{DOPE} or dense keypoints \cite{2019_Park_ICCV, 2019_Li_ICCV} of the target object and the object model to then retrieve the 6D pose via a perspective-n-point algorithm. This decouples the learning process from the fixed 6D pose representation and additionally allows the possibility to compute multiple pose hypotheses in case of uncertainty. Implicit representations based methods use the object model to render it at different poses to generate a comprehensive pose-view mapping. Later at runtime the actual pose is then determined by a match of the present view to the pose-view mapping via a latent space representation generated by an autoencoder \cite{2018_Sundermeyer_ECCV}. The advantage of this approach is, that it does not need ground truth pose annotations for training and is not suffering from pose ambiguities. 

In this context, the leading RGB-based estimator on the BOP challenge 2022 \cite{2023_Sundermeyer_CVPR} is GDR-Net \cite{Wang_2021_CVPR}, which combines and unifies the correspondence with direct approach. It even surpasses RGB-D-based estimators on the challenge, and therefore, we chose it as our RGB-based estimator. 

\subsubsection{RGB-D-based}

Leading RGB-D based approaches can be divided into deep learning approaches that are extended by classical approaches, pure classical approaches or pure deep learning approaches. The former of these make use of the additional depth information by taking an initial 6D pose of an RGB-based estimator and refine this pose by e.g. an iterative closest point (ICP) algoritm \cite{2018_Xiang_RSS, 2018_Sundermeyer_ECCV} 
or by comparing distances as in the fast version of GDRNPP \cite{GDRNPP}.
A pure classical approach, which dominated the BOP challenge until recently, is based on the matching of the observed and model point cloud via point pair features \cite{2010_Drost_CVPR}. However, a strong drawback of using depth with classical approaches is that they rely on high quality depth information and suffer from erroneous depth values, which are particularly prevalent on reflective surfaces as on the antenna. 
Against this background pure deep learning approaches process RGB and depth information jointly via a deep neural network to either estimate the 6D pose directly \cite{2019_Wang_CVPR} or via correspondences \cite{He2020_CVPR, He_2021_CVPR}. The joint learning-based approach promises to be capable to leverage the information of both modalities, while handling low-quality depth information, but still making use of it. 

For this reason, and for the reason that \cite{He2020_CVPR, He_2021_CVPR} are not straightforwardly multi-object multi-instance-per-class capable, we decided for the well established direct approach DenseFusion \cite{2019_Wang_CVPR} of this class, which additionally fulfills the requirement of a fast inference time.

\section{6D POSE ESTIMATION PIPELINE}

Following section \ref{sec:rel_work} our pipeline consists out of three state-of the-art building blocks for 6D pose estimation for rigid and known objects. One block for real-world data generation which implements LabelFusion \cite{2018_Marion_ICRA} that serves as an highly automated generator for training, evaluation and test data. One block for synthetic data generation which applies NVISII \cite{Morrical2020} and serves as a generator for unlimited synthetic training and validation data and one block for the 6D pose estimation itself. For the 6D pose estimation block we selected GDR-Net \cite{Wang_2021_CVPR} and DenseFusion (DF) \cite{2019_Wang_CVPR} as they reflect leading concepts of RGB- and RGB-D-based 6D pose estimators. To acquire the RGB and RGB-D images, we have selected the low-budget Framos D435e camera, an industrialized version of the Intel Realsense D435.  In doing so, we cover representative and well established state-of-the-art data generation and RGB- and RGB-D-based 6D pose estimation approaches.

\subsection{Real-World Data Generation}\label{rwdg}

\begin{figure}[t!]
\vspace*{1mm} % Adjust the '1cm' to suit your needs
    \centering
    \begin{overpic}[width=2.5in]{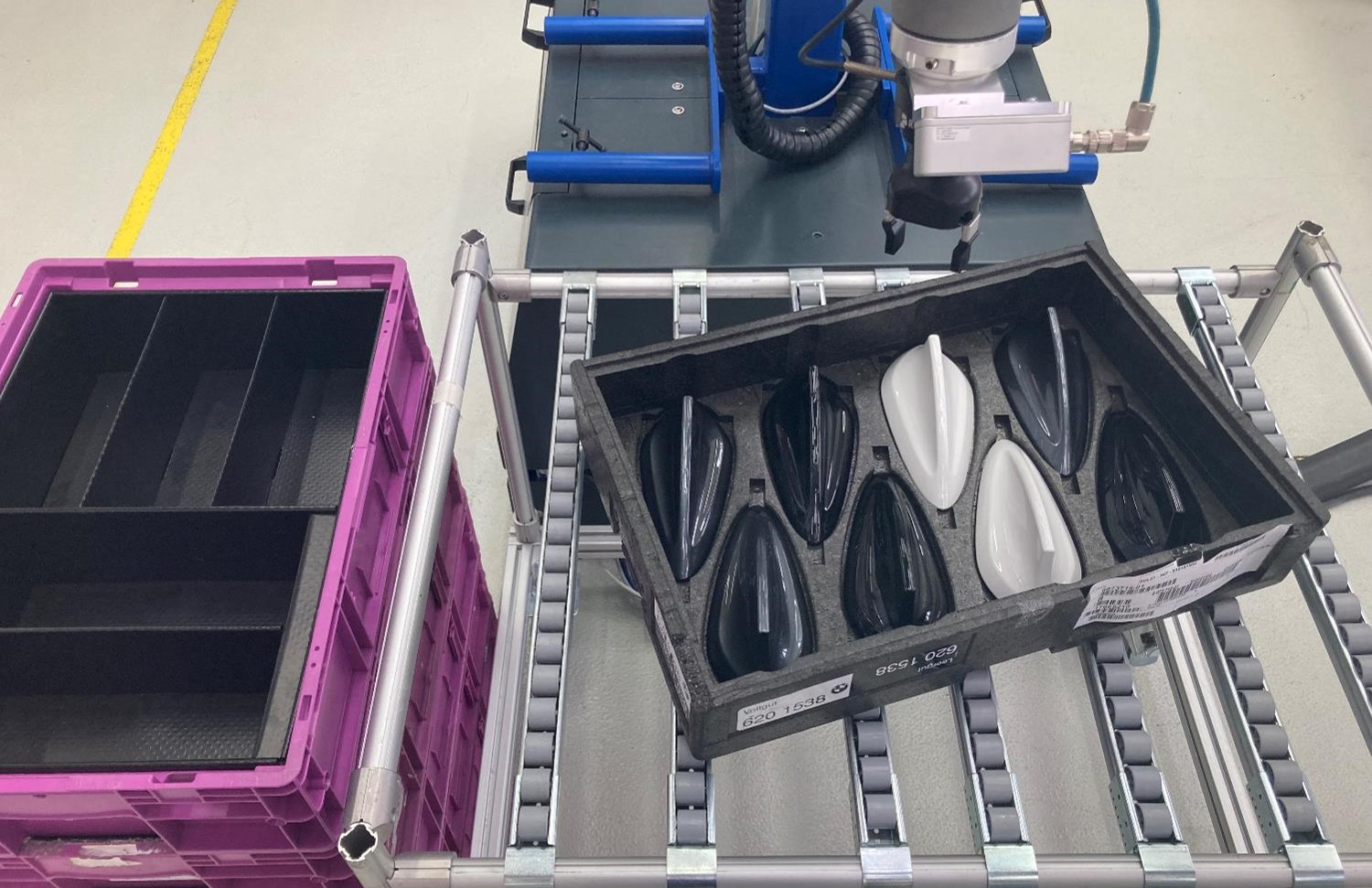}

\put (38,52) {\tikz \node[fill=white,inner sep=1pt] {\parbox{0.4in}{\centering\scriptsize\color{black}Framos\\Camera}};}

\put (55,54) {\tikz \draw[->,line width=1pt] (0,0) -- (0.6, -0.1);}

\put (7,52) {\tikz \node[fill=white,inner sep=1pt] {\parbox{0.4in}{\centering\scriptsize\color{black}Sequence\\Container}};}

\put (15,45) {\tikz \draw[->,line width=1pt] (0,0) -- (0,-0.4);}

\put (42,40) {\tikz \node[fill=white,inner sep=1pt] {\parbox{0.3in}{\centering\scriptsize\color{black}Gripper}};}

\put (55,41) {\tikz \draw[->,line width=1pt] (0,0) -- (0.5, 0.3);}

\put (88,55) {\tikz \node[fill=white,inner sep=1pt] {\parbox{0.25in}{\centering\scriptsize\color{black}UR10}};}

\put (76,56) {\tikz \draw[->,line width=1pt] (0,0) -- (-0.7, 0.4);}

\put (77,8) {\tikz \draw[->,line width=1pt] (0,0) -- (-0.1, 0.7);}

\put (70,5) {\tikz \node[fill=white,inner sep=1pt] {\parbox{0.4in}{\centering\scriptsize\color{black}Antennas}};}

    \end{overpic}
    \caption{Representative robotic setup for a automotive sequencing process showing antennas on a conveyor belt in front of a UR10 with a two finger gripper and a Framos camera mounted on the end effector. Left to the conveyor belt is the target sequence container.}
    \label{Fig:robotic_setup}
\end{figure}
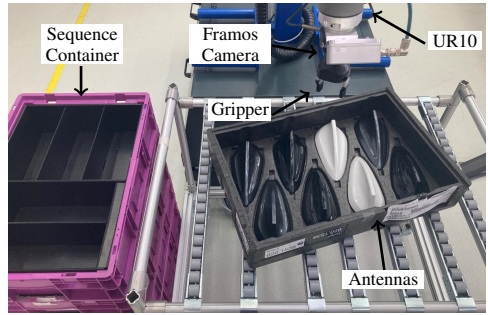

To generate annotated ground truth real-world data we use LabelFusion in combination with a partial automated scene recording. In order to create representative data we generate scenes, which are similar to sequencing for an automotive assembly line. 

In detail, we mount the camera on the end effector of a robotic arm as shown in Fig.~\ref{Fig:robotic_setup}. Then for each scene we randomly place a storage container, containing randomly arranged objects of one type, on a conveyor belt in front of the robotic-camera setup. To annotate a scene we let the camera move along a predefined trajectory to collect RGB-D samples from different viewpoints. Given these samples, LabelFusion creates a 3D reconstruction of the scene while simultaneously estimating the camera pose for each sample. In the next step the target objects in the 3D reconstruction are aligned with their corresponding models meshes through a human assisted ICP-fitting. Given the alignment, for each sample bounding boxes, masks and 6D poses are automatically retrieved, resulting in the annotation of the whole scene. 
 
This procedure enables us to quickly annotate large amounts of data. Furthermore, the usage of predefined instead of hand-held trajectories ensures reproducible results and enables cycles for iterative improvement.

\subsection{Synthetic Data Generation}\label{sec:synth_pipe}

To generate the  lightweight photorealistic scenes as described in \ref{sec:rel_synth}, we use NVISII which is an open-source PBR renderer based on Nvidia's OptiX ray tracing engine \cite{Morrical2020}.

Specifically, we create two types of such  scenes. In the first, we spawn the part's meshes in a random position on a physical floor, so that they have the same face up as in the real sequencing process. This is meant to represent simple scenes to get the model's learning process going in the right direction and to force a strong prior on this likely pose.
In the second, we spawn the part's meshes together with distractor objects in a random position above the physical floor and let them fall onto it. This way we get a scene with all kinds of poses and occlusions to further improve the model's generalization capability.

\begin{figure*}[t!]
\centering
\setlength{\fboxsep}{0pt}
\setlength{\fboxrule}{0pt}
\framebox{\parbox{7in}{
\centering
\begin{tabular}{cccc}
\includegraphics[width=0.22\linewidth]{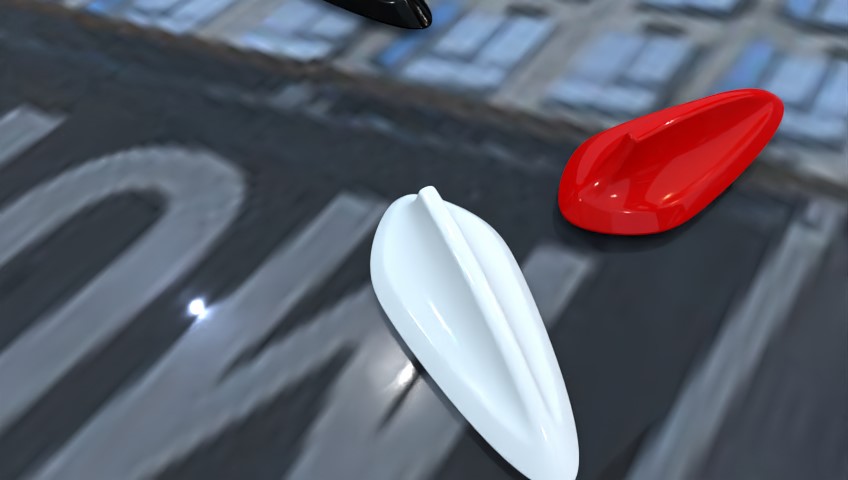} &
\includegraphics[width=0.22\linewidth]{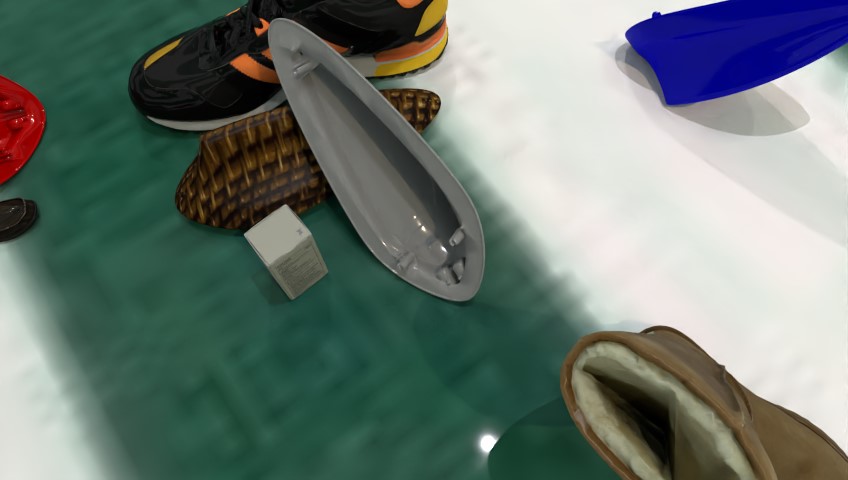}  & \includegraphics[width=0.22\linewidth]{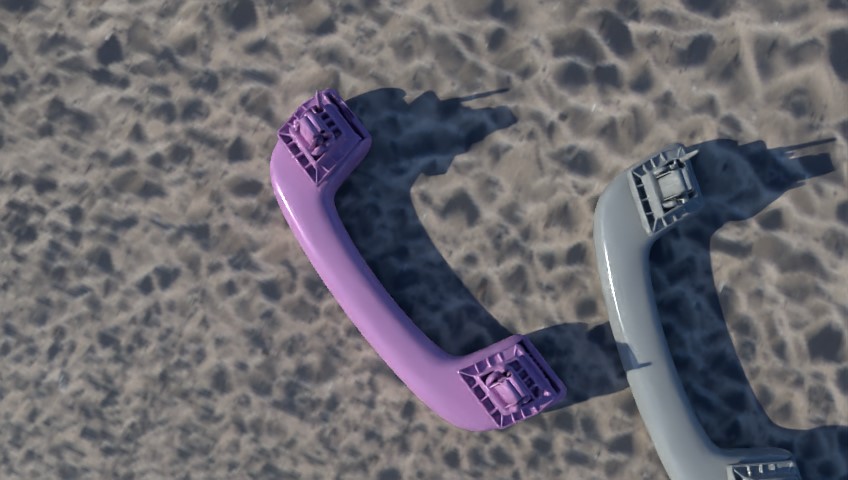} & \includegraphics[width=0.22\linewidth]{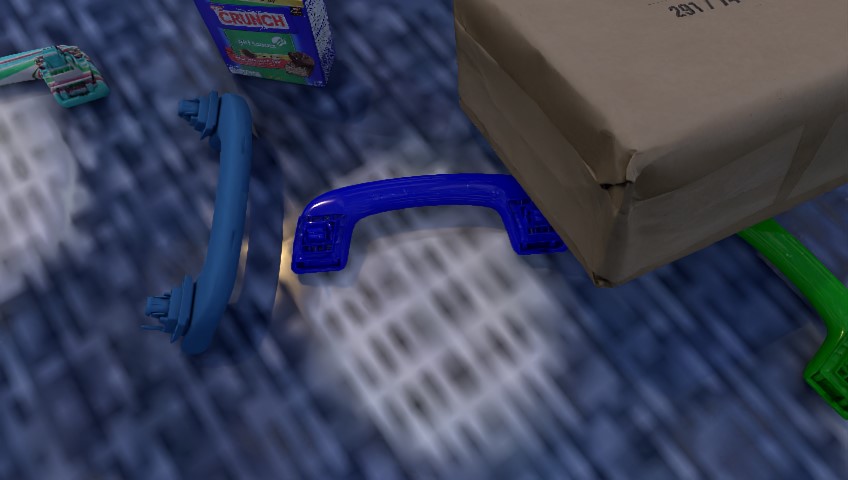} \\
\end{tabular}
}}
\caption{Exemplary rendered images showing from the left to right: First scene and second scene type antennas, first and second scene type handles.}
\label{Fig:synthetic_data}
\end{figure*}

For both scenes the parts color, texture and material properties are varied, so that we obtain a mixture of parts with a strong randomized appearance and parts with an appearance similar to that of the real world. In addition,  we randomize the floors appearance and the main scenes illumination texture with HDRI images and images from ImageNet \cite{2009_Deng_CVPR}. Besides the main illumination, we also add a random amount of spotlights at random positions. Given these scenes, RGB-D images are rendered from the perspective of a virtual camera, whose position varies randomly between different frames. Exemplary rendered images of these scenes are displayed in Fig. \ref{Fig:synthetic_data}.   
Finally, annotations in the form of ground truth poses, masks and bounding boxes are retrieved for each camera. Given this pipeline, unlimited domain randomized lightweight photorealistic data can be generated, while no additional personnel effort is needed.

\subsection{6D Pose Estimation}

For both, DF and GDR-Net we adhere to the architectures and specifications used in the original publications \cite{2019_Wang_CVPR, Wang_2021_CVPR }. For the required detection block of both estimators, we utilize Mask R-CNN \cite{2017_He_ICCV} following \cite{2021_Hofer_ICIP, 2020_Labbe_ECCV}.

\section{EXPERIMENTS}

In the experiment section, we evaluate our pipeline on representative parts of the sequencing process of an automotive assembly line. To do so, we have chosen two common parts of this process, namely antennas and handles, displayed in Fig. \ref{Fig:parts}. In this context the antenna, represents a difficult part as it is highly reflective and has no distinctive textures. The handles on the other hand, represent a rather simple part. Given these parts we applied our pipeline and conducted tests on collected data as well as on a robotic setup to examine the following questions: (1) What is the accuracy and robustness that we can reach and how does this compare to the requirements for an industrial deployment? (2) How high is the human effort, is real data still needed, or is our lightweight approach for generating synthetic data already sufficient? (3) How do the selected RGB- and RGB-D-based pose estimation approaches compare in this context?

\subsection{Metrics}

For the evaluation of the pose estimators on the data sets, we want a practical pose error function that indicates with a high likelihood the chance of a successful grasp to enable a detailed analysis with respect to process requirements.

As like the antennas and handles, the vast majority of automotive parts in the sequencing process are symmetry-ambiguity-invariant. Therefore, we follow \cite{ITODD} to use a pose error function that computes the maximum distance error $d^{P}$ (MDE) between corresponding points in homogeneous coordinates $\mathbf{x}$  of the model $\mathcal{M}$ in annotated ground truth pose $\mathbf{\bar{P}}$ and estimated pose $\mathbf{\hat{P}}$ as
\begin{equation}
d^{P}(\mathbf{\bar{P}}, \mathbf{\hat{P}} ; \mathcal{M})=\underset{\mathbf{x} \in \mathcal{M}}{\operatorname{max}}\|\mathbf{\bar{P}} \mathbf{x}-\mathbf{\hat{P}} \mathbf{x}\|_{2}
\end{equation} 
where a pose is defined as a homogeneous transformation matrix $\mathbf{P}=[\mathbf{R} \mid \mathbf{t}]$ $\in$ SE(3), consisting of a rotation matrix $\mathbf{R} \in$ SO(3) and a translation vector $\mathbf{t} \in \mathbb{R}^{3}$. In doing so, we enable a practical but conservative determination of a single threshold towards a successful grasp for a wide range of grippers.  

Based on this, we can calculate performance scores for industrial requirements. In our case we must avoid false positives (FP), as they can lead to robotic crashes and damaged parts. Therefore, we are interested in the precision score. Furthermore, we need also take measure on false negatives (FNs) as they lead to longer process times and can make a pose estimator impractical. Thus, we are also interested in the recall score.   
To this end we calculate the average precision (AP) and average recall (AR) for $N$ samples as 
\begin{equation}
AP = \frac{1}{N} \sum_{n=1}^{N} \frac{TP_{n}}{TP_{n}+FP_{n}} \label{eq:ap}
\end{equation}
\begin{equation}
AR =  \frac{1}{N} \sum_{n=1}^{N} \frac{TP_{n}}{TP_{n}+FN_{n}} \label{eq:ar}
\end{equation}
where $TP_{n},FP_{n} $ and $FN_{n}$ are the number of the true positives (TPs), FPs and FNs in a sample $n$, respectively.

To compute these scores we match all pose estimations of a sample $n$ to the ground truth (GT) annotations in increasing order of the corresponding maximum distance error $d^{P}$, while every estimation can only be matched once but every GT multiple times. All matched estimations whose $d^{P}$ error is smaller or equal to an error threshold $\theta_{p} $ are counted as a TP and a FP  otherwise. From all GTs, only GTs that do not have a TP match and have a visibility ratio $v$ of a corresponding object in a sample greater or equal to a visibility threshold $\theta_{v} $, are counted as FN. The reasoning behind this is that in our case we do not have partial visibility due to occlusion but only due to process dependent field of view (FoV) restrictions. Therefore, we do not want to penalize missing estimations for objects, for which it is not intended to predict poses for, as the visible part of such objects may not even contain enough information for an optimal 6D pose estimator. Nevertheless, if the estimator provides a pose for such an object, its error has to be smaller or equal to the error threshold otherwise it is counted as a FP. In this context we set the $\theta_{v} = 0.85 $, as we require from the process to enable such a visibility and the error threshold  
$\theta_{p} = 1.5$ cm, as this should be the upper limit for a successful grasp for a wide variety of part-gripper combinations. 

\subsection{Datasets}

\subsubsection{Real-World Data}
Using our real-world data pipeline we generated a data set for each part. We collected the RGB-D images with a resolution of 848x480 pixels, which is the resolution of the Framos to achieve the highest depth quality. For each part we decided to collect 30 scenes at the heights around of 30 cm, 10 scenes 50 cm  and 10 scenes 80 cm above the storage container. The height of 30 cm is intended to serve as a distance from which the 6D pose for the grasping will be estimated and therefore makes up the majority of the collected scenes. This height is the minimal distance at which the complete storage container is still in the FoV and provides a sufficient  margin for successful depth images acquisition. The heights of 50 and 80 cm are intended to make the estimators robust against height variations. We divided the total of 50 scenes for each part into 30 training scenes with about 80k images, 10 validation and 10 test scenes with about 25k images each. Each dataset could be generated by one person in about 4 days.

\subsubsection{Synthetic Data} 

Using our synthetic data pipeline, we generated 25,000  annotated training and 5,000 validation images for each scene type and part. This results in a total of 50,000 training and 10,000 validation images for each part, which is very similar to the seizes in \cite{DOPE, Hodan2020}. The distance of the camera to the target parts was chosen to be similar to the real data, with a distance between 30 and 80 cm. The scenes themselves contain randomly up to 10 distractors, spotlights and target parts each. Example images of the data sets are shown in Fig. \ref{Fig:synthetic_data}.

\subsubsection{Dataset Variants} 

In order to evaluate our research questions, we created a real world (R), a synthetic (S) and a combined data set (R-S) for each part as specified in Tab. \ref{tab1}. 
\begin{table}[t]
\caption{Dataset Variants per Part}
\begin{center}
\begin{tabular}{|c|c|c|c|}
\hline
\textbf{Name} & \textbf{Train}& \textbf{Validation}& \textbf{Test} \\
\hline
R & 80k real img. & 25k real img.  &  \\ 

   \cline{1-3}

S  & 50k synth. img & 10k synth. img. & 25k real img. \\

         \cline{1-3}
R-S & R Train + S Train & R Val. + S Val.&    \\

\hline
\end{tabular}
\label{tab1}
\end{center}
\end{table}

\subsection{Implementation and Training Details}

To train Mask R-CNN, we use a pre-trained ResNet-50. For the training of DF, we used the same training hyperparameters as in the original publication with 2 refinement iterations and a refinement threshold of 8 mm, but with a pre-trained instead of an untrained ResNet-18. For GDR-Net, we also used the same hyperparameters as in the original publication, but excluded the built-in data augmentation, as we want to compare both estimators on the same data set.   

Furthermore, for both 6D pose estimators, we excluded annotations in the training and validation parts of the data sets of objects that do not exceed a visibility of $\theta_{v} = 0.3$, as we observed that lower thresholds worsen the results. Moreover, we trained and validated on the GT instead on inferred detections, as we observed better results when testing such models with inferred detections on the test set.

\subsection{Results on Dataset Variants}

\subsubsection{Instance Segmentation} 
The results for the trained Mask R-CNNs evaluated on the test set are shown in Table \ref{tab2}. The results are given in percentage for $AP_{50(\%)}$ (the average precision at IoU 50 \%), $AP_{50:95(\%)}$ and $AR^{100}$ (the average recall for at maximum 100 detections per image).
\begin{table}[t]
\caption{Mask R-CNN Results }
\begin{center}
\begin{tabular}{|c|c|c|c|c|}
\hline
\textbf{Object} & \textbf{Dataset}& \textbf{$\mathbf{AP_{50:95(\%)}}$}& $\mathbf{AP_{50(\%)}}$ & \textbf{$\mathbf{AR^{100}}$}  \\
\hline
& R & 85.2 & 99.0 & 89.0  \\ 
Antenna   & S & 39.5 & 55.8 & 44.1 \\
    & R-S & 85.7 &  99.0 & 89.8 \\
   \cline{1-5}
& R & 73.5 & 99.0 & 78.5 \\ 

Handle     & S & 63.8 & 96.1 &  68.5 \\
& R-S & 75.3 & 98.9 & 78.9 \\
  
\hline
\end{tabular}
\label{tab2}
\end{center}
\end{table}
It can be concluded that both the models trained on real data and the model trained on real and synthetic data combined reach state of the art scores.
The model trained on synthetic data only, do not, especially the version for the antenna.

\subsubsection{6D Pose Estimation} 
For the evaluation of the trained 6D pose estimation models, we used the inferred  detections from the Mask R-CNNs of the same data variant. Furthermore, we used the possibility that DF provides confidences $c \in [0,1]$ \cite{2019_Wang_CVPR} as an uncertainty measure for its estimation, whereas GDR-Net does not. Unless otherwise specified, we have chosen the confidences for DF so that the AP (\ref{eq:ap}) is maximized, but at maximum to $c = 0.95 $, as we want to prevent robotic crashes. The confidences for the DF models for the Antenna/Handle models were set to R: 0.95/0.8, S: 0.5/0.7, R-S: 0.95/0.95.

In Tab. \ref{tab5} we tabulate the AP (\ref{eq:ap}) and AR (\ref{eq:ar}) scores  in  \% for the different DF and GDR-Net models.
\begin{table}[tb]
\caption{Avg. Precision \& Avg. Recall Dataset Results}
\begin{center}

\begin{tabular}{|c|c|c|c|c|c|c|c|}
\hline
\textbf{Object} & \textbf{Method}& \multicolumn{2}{|c|}{\textbf{R}}&  \multicolumn{2}{|c|}{\textbf{S}}  &  \multicolumn{2}{|c|}{\textbf{R-S}} \\
 \cline{1-8}
  & & AP & AR & AP & AR & AP & AR   \\
   \cline{3-8}
  
Antenna & \begin{tabular}{@{}c@{}}DF \\ GDR\end{tabular}  & \begin{tabular}{@{}c@{}} 95.2 \\ 77.4 \end{tabular} & \begin{tabular}{@{}c@{}} 89.3 \\ 96.3 \end{tabular} & \begin{tabular}{@{}c@{}} 70.8 \\ 73.0 \end{tabular} & \begin{tabular}{@{}c@{}} 28.7 \\ 40.9 \end{tabular} & \begin{tabular}{@{}c@{}} 96.6 \\ 78.6 \end{tabular} & \begin{tabular}{@{}c@{}} 92.7 \\ 97.2 \end{tabular} \\
\cline{1-8}
Handle  & \begin{tabular}{@{}c@{}}DF \\ GDR\end{tabular}  & \begin{tabular}{@{}c@{}} 97.7 \\ 82.6 \end{tabular} & \begin{tabular}{@{}c@{}} 91.6 \\ 95.3 \end{tabular} & \begin{tabular}{@{}c@{}} 96.3 \\ 29.4 \end{tabular} & \begin{tabular}{@{}c@{}} 41.5\\ 37.3 \end{tabular} & \begin{tabular}{@{}c@{}}99.5 \\ 86.6 \end{tabular} & \begin{tabular}{@{}c@{}} 92.8 \\ 98.4 \end{tabular}  \\
\hline
\end{tabular}
\end{center}
\label{tab5}

\end{table}
The result show that especially the DF models that included real data in the training are close to an optimal AP of 1 and perform in that term better than the GDR-Net counterpart. Nevertheless, despite that DF provides confidences as an uncertainty measure to adapt the AP there is no model that reaches an optimal AP. Additionally, it can be seen that the GDR-Net predominantly achieves higher AR values, which was to be expected since the poses were not filtered by an uncertainty measure. But again, despite the visibility constraint, optimal AR values are not achieved. Remarkable is the near optimal AP of the DF model trained only with synthetic data on the handle and the generally high scores of the synthetic models with the exception of the GDR-Net on the handle.

\begin{figure}[tb]
\centering
\setlength{\fboxsep}{0pt}
\setlength{\fboxrule}{0pt}
\framebox{\parbox{3in}{\centerline{\includegraphics[width=0.49\textwidth]{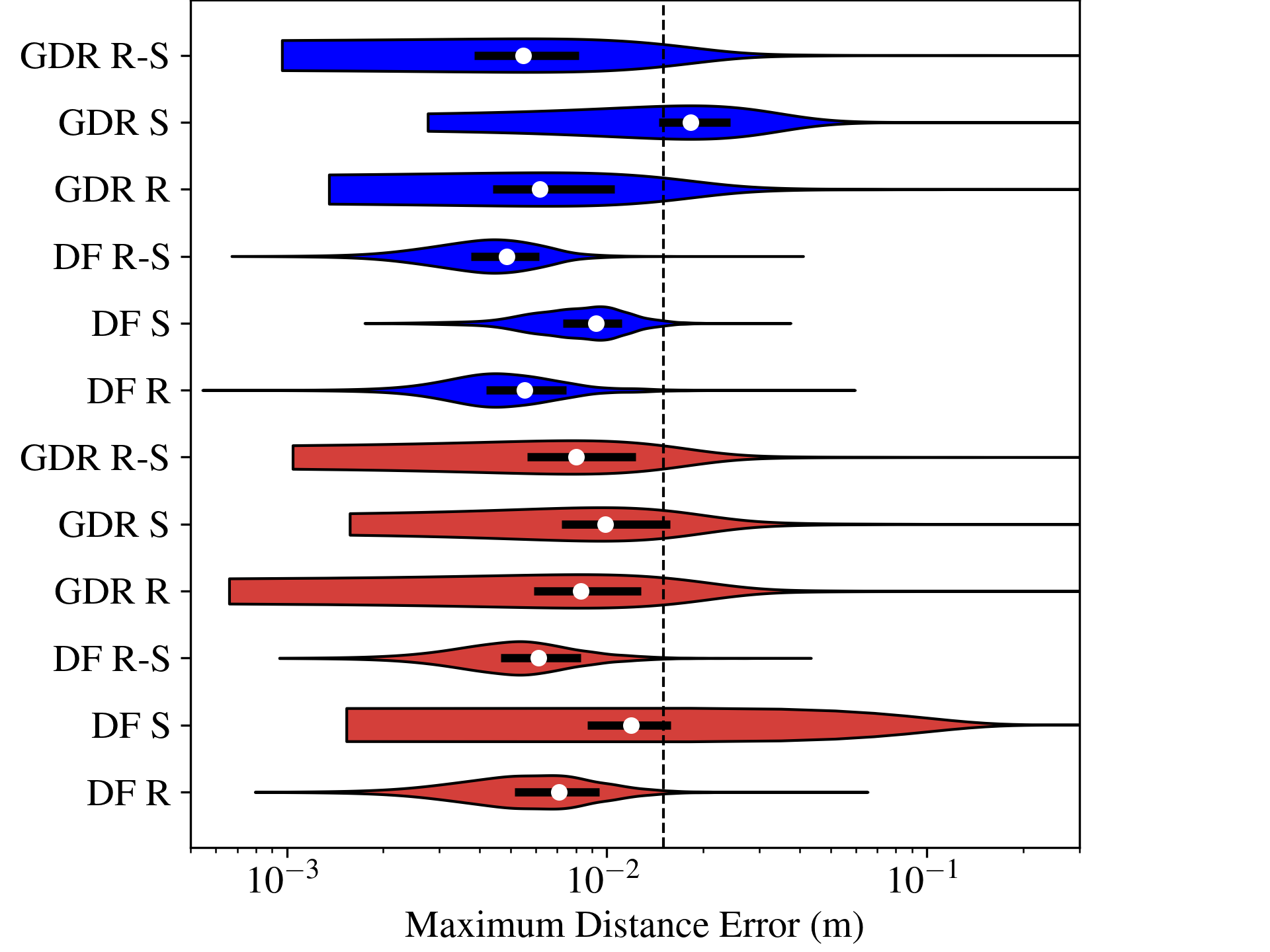}}
}}    
\caption{Violine plots of the MDE on the test set for the different models on a logarithmic axis. The results for the antenna are shown in red and for the handle in blue. The white dots represent the location of the median and the black bars indicate the location of the lower and upper quartiles, enclosing 75 \% of the data. The vertical dotted line shows the error threshold $\theta_{p} $ of 0.015 m. The x-axis is limited from 0.5 mm to 0.3 m.}
\label{violinplot}
\end{figure}

To additionally analyze the distribution of the models MDE's we complement the AP and AR scores with a violin plot in Fig. \ref{violinplot}.
The plot shows that most models have the majority of their MDEs below the critical threshold of $\theta_{p} = $ 0.015 m. Furthermore, it is noteworthy that for the antenna the synthetic model of GDR-Net and for the handle the synthetic model of DF, have similar MDE distributions compared to their real counterparts. However, it is noticeable that all models have quite high MDE outliers and even the best performing model has outliers up to 4 cm.

Furthermore, we tabulate the mean and standard deviation of the x-,y- and z-error components of the ADD error in mm \cite{2012_Hinterstoisser_ACCV} in Tab. \ref{tab:component_errors} to enable an analysis of the error sources.  
\begin{table}[tb]
\caption{ADD Error Components}
\begin{center}
\begin{tabular}{|cc|cc|cc|cc|cc|}
\hline
 & & \multicolumn{4}{|c|}{Antenna}&  \multicolumn{4}{|c|}{Handle}   \\
\cline{3-10}
  &     &  \multicolumn{2}{|c|}{DF} & \multicolumn{2}{|c|}{GDR}  &  \multicolumn{2}{|c|}{DF} & \multicolumn{2}{|c|}{GDR} \\
 
   &     &  $\overline{x}$ & $s$  &  $\overline{x}$ & $s$ &  $\overline{x}$ & $s$ &  $\overline{x}$ & $s$\\
\cline{1-10}
R & \begin{tabular}{@{}c@{}@{}}x \\ y \\ z\end{tabular} & \begin{tabular}{@{}c@{}@{}} 2.41  \\ 2.81  \\ 3.10   \end{tabular} & \begin{tabular}{@{}c@{}@{}} 1.73 \\ 2.21 \\ 2.43  \end{tabular} & \begin{tabular}{@{}c@{}@{}}7.57  \\ 6.03  \\ 16.4  \end{tabular} & \begin{tabular}{@{}c@{}@{}}27.4  \\19.3 \\ 64.1  \end{tabular}  & \begin{tabular}{@{}c@{}@{}}2.41  \\ 2.12   \\ 2.04   \end{tabular} & \begin{tabular}{@{}c@{}@{}} 2.40  \\ 1.95 \\ 1.23 \end{tabular} & \begin{tabular}{@{}c@{}@{}}7.70  \\ 6.04  \\ 14.2  \end{tabular} & \begin{tabular}{@{}c@{}@{}}23.5  \\21.2  \\ 51.9  \end{tabular}\\
\cline{1-10}
S & \begin{tabular}{@{}c@{}@{}}x \\ y \\ z\end{tabular} & \begin{tabular}{@{}c@{}@{}}9.45  \\ 7.21  \\ 16.3   \end{tabular} & \begin{tabular}{@{}c@{}@{}} 134 \\ 64.2  \\ 385  \end{tabular} & \begin{tabular}{@{}c@{}@{}} 5.02 \\ 4.93 \\ 14.0  \end{tabular} & \begin{tabular}{@{}c@{}@{}} 12.9  \\ 11.3  \\ 41.5  \end{tabular} & \begin{tabular}{@{}c@{}@{}}1.84 \\ 1.74  \\ 4.75   \end{tabular} & \begin{tabular}{@{}c@{}@{}} 1.41  \\ 1.10  \\ 1.82  \end{tabular} & \begin{tabular}{@{}c@{}@{}} 7.20  \\ 7.22  \\ 24.9  \end{tabular} & \begin{tabular}{@{}c@{}@{}} 21.7  \\ 25.5  \\ 77.7  \end{tabular} \\
\hline
\end{tabular}
\label{tab:component_errors}
\end{center}
\end{table}
Note that in this table a comparison between the real and synthetic data models need to be done with caution, as they had a different detection input and are therefore not evaluated on the same set of instances. 

\begin{figure*}[tb]
\begin{center}
\begin{tabular}{cc}
\includegraphics[width=0.4\linewidth]{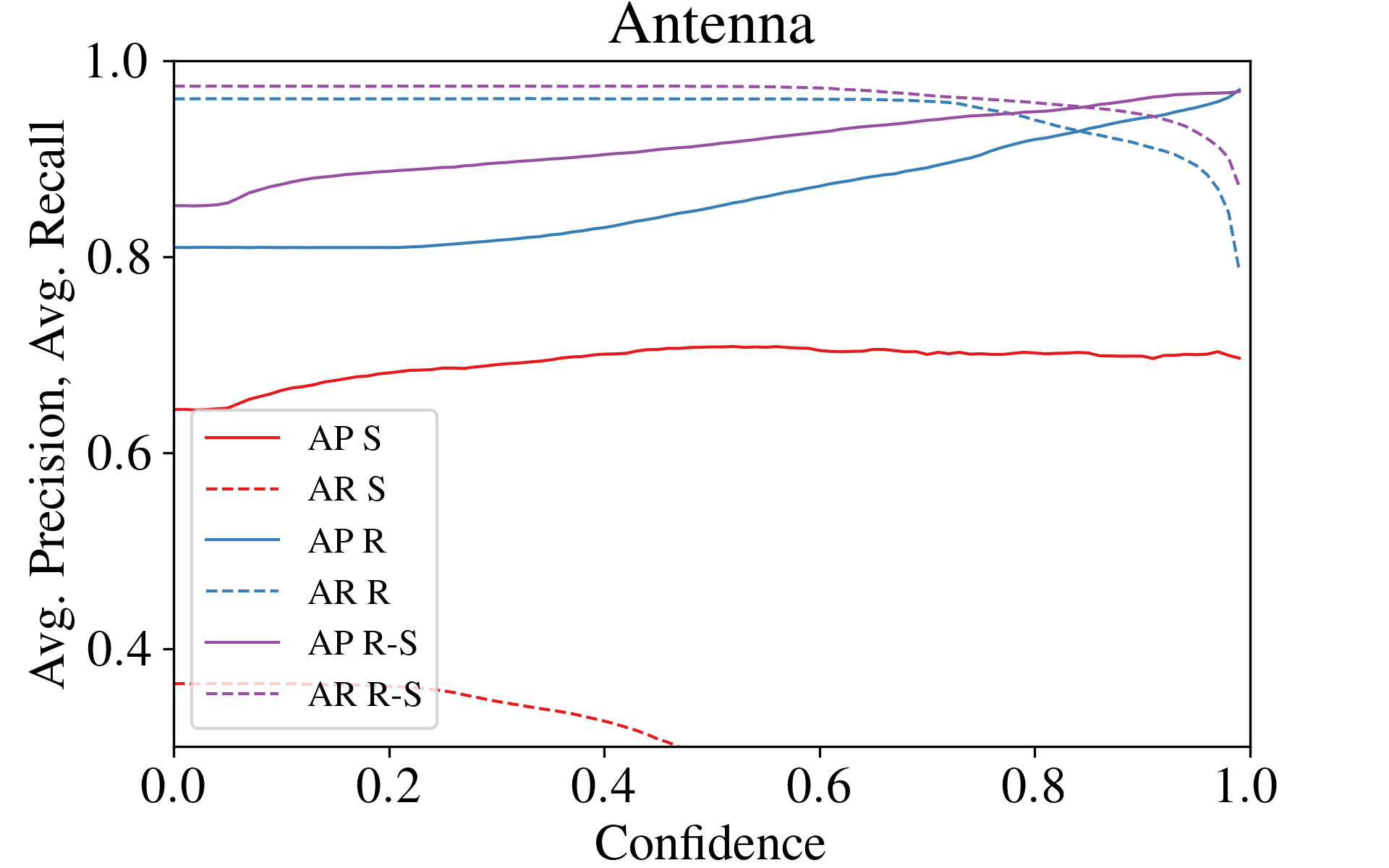} &
\includegraphics[width=0.4\linewidth]{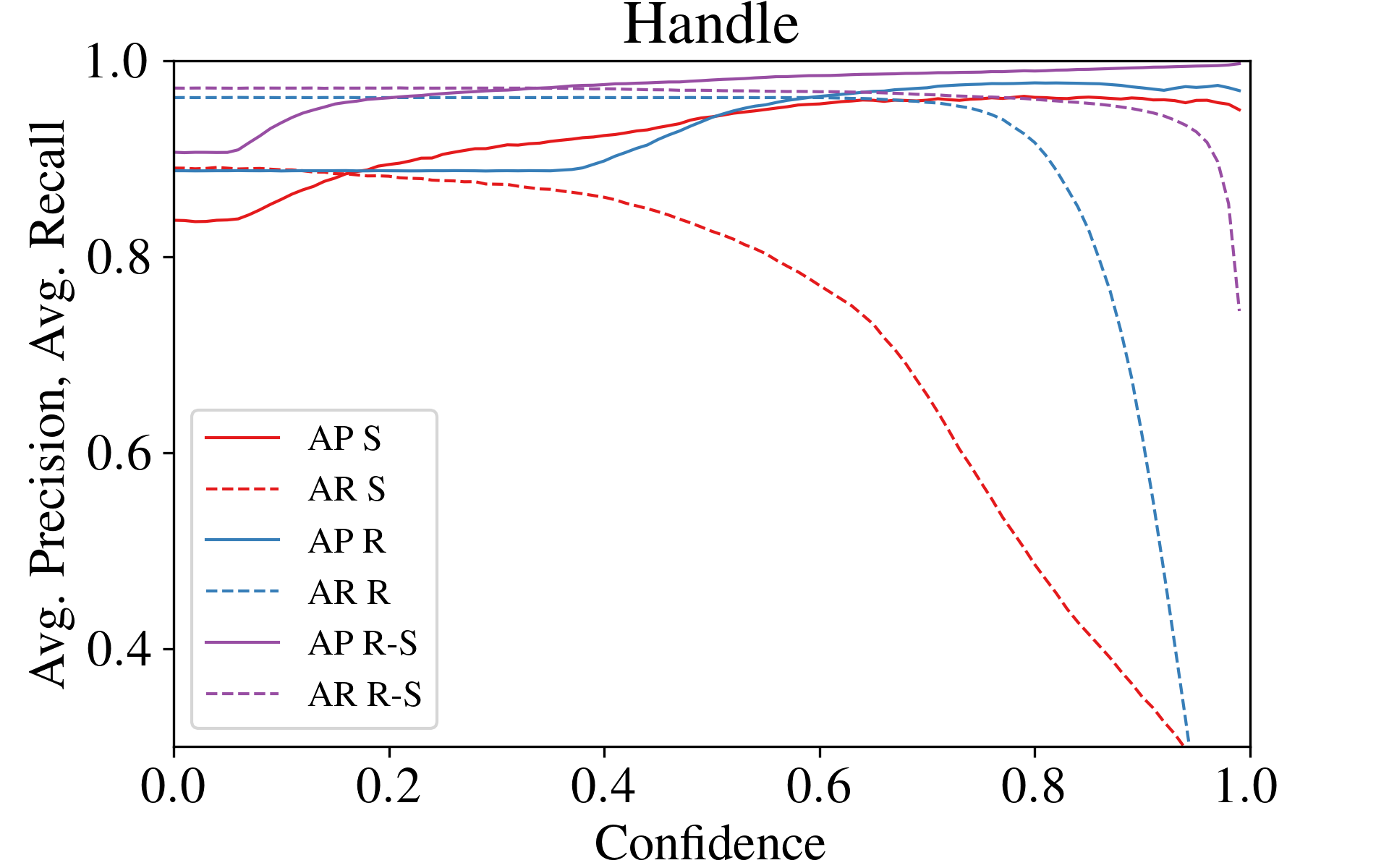} \\
\end{tabular}
\end{center}
\caption{Average Precision and Average Recall of DF over a confidence intervall $c  \in [0,1)$  for the Antenna on the left and the Handle on the right.}
\label{Fig:Graphs}
\end{figure*}

To additionally enable a more in depth analysis of the confidence variable of DF, we display the AP and AR curves over a confidence interval $c  \in [0,1)$ for each DF model in Fig. \ref{Fig:Graphs}.   
The curves show that the models have a low sensitivity of the AR and AP scores with respect to the confidence. In particular, the AP course has mostly constant or occasionally even lower scores for higher confidence values and no optimal AP can be set. This problem is particularly evident for the models trained on synthetic data.
  
\subsection{Experiments on Robotic Grasping Setup}
To evaluate how the models would perform in a real sequencing process, we performed tests on the robotic setup of Fig. \ref{Fig:robotic_setup} with a two-finger gripper that is capable of grasping both parts, and is relatively robust to pose inaccuracies. To conduct the experiments, we placed a randomly oriented storage container containing randomly arranged parts of one type on a conveyor belt in front of the robotic setup.  The goal was then to sequentially pick all parts in the storage container and place them in a nearby sequence container, where for each attempt a new image of the scene was acquired. To do this, the robot performed a search movement so that the entire area of the conveyor belt was successively within the FoV of the camera. To have equal conditions, each model was tested on the same set of 5 randomly prepared scenes per part, where each scene contained 8 parts.

As the experiments offer the opportunity to make the AP and AR scores more meaningful for evaluating a deployment, we deviated from the definition of TP, FP and FN of the data set evaluation. Specifically, we counted an attempt as a TP if the robot managed to place a part in the sequence container. The slot position in this container was hard-coded, and to account for acceptable inaccuracies, the slot was chosen to be 20 mm larger in x and y directions than the maximum x and y dimensions of the respective parts in the intended placement position. An attempt was counted as a FP, if it lead to a crash or to an unsuccessful placement. We removed each part after it caused a crash so a FP could only be counted once per part. A FN was counted for each part that remained in the container after 160 seconds, which equals 20 seconds per part and is a realistic assumption. The parts that were removed because of a crash were counted as a FN. 

We display the results of the experiment in Table \ref{tab6}.  
\begin{table}[tb]
\caption{Avg. Precision \& Avg. Recall Robotic Experiment Results}
\begin{center}
\begin{tabular}{|c|c|c|c|c|c|c|c|}
\hline
\textbf{Object} & \textbf{Method}& \multicolumn{2}{|c|}{\textbf{R}}&  \multicolumn{2}{|c|}{\textbf{S}}  &  \multicolumn{2}{|c|}{\textbf{R-S}} \\
 \cline{1-8}
  & & AP & AR & AP & AR & AP & AR   \\
   \cline{3-8}
  
Antenna & \begin{tabular}{@{}c@{}}DF \\ GDR\end{tabular}  & \begin{tabular}{@{}c@{}} 100 \\ 100 \end{tabular} & \begin{tabular}{@{}c@{}} 87.5 \\ 100 \end{tabular} & \begin{tabular}{@{}c@{}} 60.8 \\ 87.1 \end{tabular} & \begin{tabular}{@{}c@{}} 52.5 \\ 85.0 \end{tabular} & \begin{tabular}{@{}c@{}} 100 \\ 97.5 \end{tabular} & \begin{tabular}{@{}c@{}} 100 \\ 97.5 \end{tabular} \\
\cline{1-8}
Handle  & \begin{tabular}{@{}c@{}}DF \\ GDR\end{tabular}  & \begin{tabular}{@{}c@{}} 100 \\ 97.5 \end{tabular} & \begin{tabular}{@{}c@{}} 97.5 \\ 87.5 \end{tabular} & \begin{tabular}{@{}c@{}} 97.5 \\ 0.00\end{tabular} & \begin{tabular}{@{}c@{}} 92.5 \\ 0.00 \end{tabular} & \begin{tabular}{@{}c@{}} 100 \\ 80.0\end{tabular} & \begin{tabular}{@{}c@{}} 95.0 \\ 60.0\end{tabular}  \\
\hline
\end{tabular}
\label{tab6}
\end{center}
\end{table}
The results show that optimal AR and AP values could be obtained for both parts with DF and GDR-Net. It should be emphasized that the purely synthetically trained DF model has nearly optimal values for the handles. In contrast, it performs relatively poorly for the antennas. We observed that this was especially the case when the depth images were erroneous. Fig. 6 shows such an example from the experiment. In contrast, GDR-Net trained on synthetic data was much more robust on the antennas. However, the poor values of the synthetically trained GDR-Net for the handle are particularly striking. Here, we observed that the poses were generally estimated too high and the grab was reaching into the void. In this context Tab. \ref{tab:component_errors} shows that the ADD error on the z-coordinate for this synthetic model is 24.9 mm compared to 14.2 mm for the real counterpart, while the errors on the x- and y-coordinate are in the same range. 
\begin{figure}[tb]
\begin{center}
\begin{tabular}{c}
	\begin{overpic}[width=2.7in]{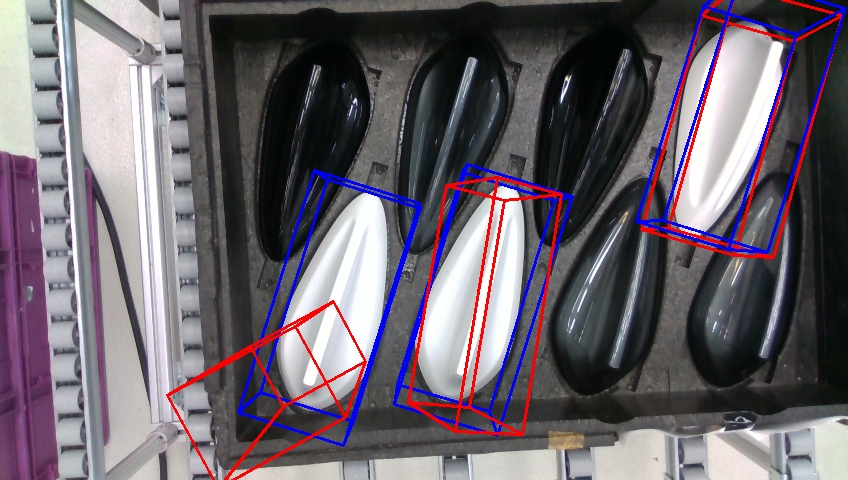}
	\end{overpic}\\
	\begin{overpic}[width=2.8in]{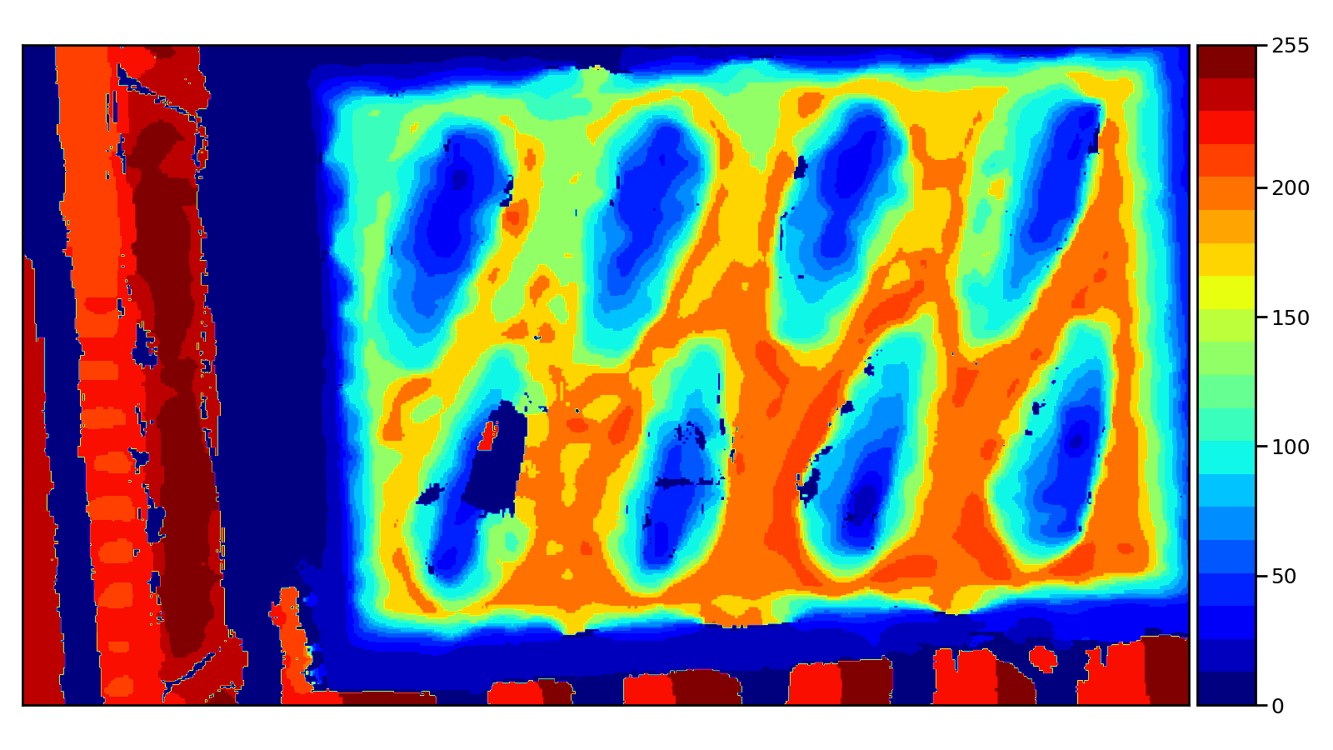}
	\put (10,23) {\tikz \node[fill=white,inner sep=1pt] {\parbox{0.4in}{\centering\scriptsize\color{black}Erroneous\\Depth}};}

\put (25,24) {\tikz \draw[->,line width=1pt] (0,0) -- (0.7, -0.2);}
	\end{overpic}
\end{tabular}
\end{center}
\caption{RGB image with projected pose estimations of DF (red) and GDR (blue) in the form of 3D bounding boxes with a scaled heatmap of the corresponding depth image below. The wrongly estimated poses of DF on the bottom left correspond to depth areas which have false depth values.}
\label{Fig:heatmap}
\end{figure}

\section{DISCUSSION}

Towards research question (1) we state that despite the existing nearly optimal results in the robotic experiments for both parts, it is questionable that these numbers would hold for continuous industry operation, as no model in the bigger data set achieved such results and Fig. \ref{violinplot} shows high outliers.
In particular, we conclude that industry requirements are not met, especially in terms of robustness rather than accuracy. The biggest issue in this context is the guarantee of an optimal AP, as a measure of robustness, which, if not adhered to, leads to robot crashes with damaged parts and downtime. In our view, a major reason for not achieving the optimal AP, is the inability of the models to provide a reliable uncertainty measure rather than the ability to provide sufficiently accurate poses. Fig. \ref{violinplot} in particular shows that the majority of the estimated poses have a sufficiently small error. This is also true for most models that have been trained with synthetic data only. In this connection, DF provides a confidence as an uncertainty measure, but the results show that its properties are not sufficient to reliably meet an AP of 1. GDR-Net, although leading the BOP challenge 2022 \cite{2023_Sundermeyer_CVPR}, is even worse, as it does not provide any uncertainty at all.  Without such, we conclude that a deployment is not robust to unavoidable occlusions or erroneous sensor data.

However, independent of this issue we can state towards research question (2) that the data set generation effort-benefit relationship, is very promising in terms of economic scalability. By using our pipeline, the real data sets could be generated by one person in about 4 days each and with our lightweight synthetic data approach, no further personnel effort is needed. The results show that there is reasonable evidence to assume that given an reliable uncertainty estimation, the results achievable with the real data are sufficient for an industrial deployment. 
Regarding the question if this would also hold for the lightweight synthetic data approach, we found that, despite the promising results, it depends on the used estimator-data-modality combination (3).

In this context the example of the reflective antenna shows that erroneous depth data leads to a large domain gap with wrong poses, for the DF as the representative of an RGB-D based approach. The GDR-Net, as a representative of an RGB-based approach, that does not use such depth data, performed significantly better in this case. On the other hand, the missing depth makes the GDR-Net generally more inaccurate on the distance estimate, which is likely to lead to insufficient results, if the gripping point is intolerant against such errors, which was the case for the handle in combination with synthetic data. We assume that the domain gap amplifies the disadvantage of an RGB-based estimator of not using direct depth information.
Against this background, we suggest that the lightweight approach can be sufficient with either an RGB-D or RGB-based approach for parts that are not reflective, or if they are, a grab point must be available that is tolerant of z-errors.

Note that the generalizability of the conclusion drawn on our research question may be limited by the carefully selected parts and estimators. Nevertheless, we are convinced that the results provide valid insights into these questions.

\section{CONCLUSIONS}

In this work, we evaluated a representative 6D pose estimation pipeline to investigate the status quo of whether automotive internal logistics can be competitively automated by robots. In this context, we show that a major problem to enable such a competitive automation is that the estimators do not provide reliable uncertainties for their estimates. Therefore, we suggest a strong focus of further research on this property. Furthermore, we show that the selected representative RGB approach is more robust than the selected representative RGB-D approach when a domain gap is caused by erroneous depth data. Since RGB is inherent to the RGB-D approach, this implies that RGB-D approaches can be significantly improved.
This could be achieved by improving the network's awareness of erroneous depth data, or by incorporating such errors into the synthetic data generation and training of the networks. Finally, this work demonstrates the difficulties in transferring the promising scientific results of 6D pose estimation to industrial use cases.

\section*{ACKNOWLEDGMENT}

The authors would like to thank Matthias Brucker and Luca Della Libera for their help on refactoring DF and LabelFusion, Jan-Oliver Seidenfuss and Daniel Derkacz-Bogner for their help on the synthetic data pipeline.

%\addtolength{\textheight}{-12cm}   % This command serves to balance the column lengths
                                  % on the last page of the document manually. It shortens
                                  % the textheight of the last page by a suitable amount.
                                  % This command does not take effect until the next page
                                  % so it should come on the page before the last. Make
                                  % sure that you do not shorten the textheight too much.

%%%%%%%%%%%%%%%%%%%%%%%%%%%%%%%%%%%%%%%%%%%%%%%%%%%%%%%%%%%%%%%%%%%%%%%%%%%%%%%%

%%%%%%%%%%%%%%%%%%%%%%%%%%%%%%%%%%%%%%%%%%%%%%%%%%%%%%%%%%%%%%%%%%%%%%%%%%%%%%%%

%%%%%%%%%%%%%%%%%%%%%%%%%%%%%%%%%%%%%%%%%%%%%%%%%%%%%%%%%%%%%%%%%%%%%%%%%%%%%%%%

\bibliographystyle{IEEEtran}
\bibliography{IEEEabrv,../../jabref/dissertation.bib}

\end{document}